\documentclass[runningheads]{llncs}
\usepackage[T1]{fontenc}
\usepackage{listings}
\usepackage{xcolor}
\usepackage{float}
\newfloat{listing}{htbp}{lol} 
\floatname{listing}{Listing}
\usepackage{todonotes}

\usepackage{graphicx}
\begin{document}
\title{A New Approach to Characterising Optimisation Problems Using Programmatic Representation and Complexity Measures}
%
\titlerunning{Characterising Problems Using Programmatic Representation}
%
\author{Marcus Gallagher\inst{1}\orcidID{0000-0002-6694-9572} \and\\
Katherine M. Malan \inst{2}\orcidID{0000-0002-6070-2632}}
\authorrunning{Gallagher and Malan}
%
\institute{School of Electrical Engineering and Computer Science,\\ University of Queensland, Australia. 
\email{marcusg@uq.edu.au}
\and
Department of Decision Sciences, University of South Africa, Pretoria, South Africa \email{malankm@unisa.ac.za}}%

\maketitle              
\begin{abstract}
Characterising optimisation problem instances is a fundamental part of understanding the behaviour and performance of different algorithms as well as providing information for algorithm selection and configuration. In this paper we propose a novel approach to problem characterisation based on the representation of instances when implemented as a program. The intuition is that the complexity of the code required to express an objective function should relate to the complexity of the search landscape. We identify the Halstead volume as a 
measure of code complexity, which can be seen as a simplified version of the entropy of the program. Given a code implementation of the objective function, the Halstead volume and entropy can be quickly calculated using existing libraries.
We apply the proposed complexity measures to the well-known BBOB optimisation problem suite and the simple feed-forward neural network training task. We also show that the measures are negatively correlated with algorithm performance and therefore show potential as predictive meta-features for algorithm selection and other problem analysis. We envisage the proposed measures as complementary to other problem characterisation approaches, but with the advantages of not requiring any sampling of the search space, being invariant to transformations, and being very quick to calculate automatically.

\end{abstract}
\section{Introduction}
In metaheuristic optimisation, the performance of an algorithm depends on both the operators and internal mechanisms of the algorithm and the properties of the problem instance(s) on which it is applied. To solve new problems effectively, we need to be able to understand and identify the relationship between algorithm performance and problem properties. This requires the ability to characterise or distinguish between problem instances. This is an active area of research, where different approaches have been taken (e.g. using information available from the formulation or representation of the problem, or by using information gained by exploring the landscape). See Section~\ref{sec:litrev} for a more detailed overview. 

In a general sense, any summary description of an optimisation problem must incur a loss of information. Practical assumptions and limitations are also important, e.g. in exploratory landscape analysis the computational effort required to generate/evaluate a sample of candidate solutions and the modelling assumptions behind landscape features. As a result, it is likely that different classifications/categorisations/features of problem instances will provide different information or representations. While one specific problem feature might be, for example, highly predictive of the performance of a specific algorithm, different types of features in combination may provide complementary information to more completely characterise a problem instance.

This paper proposes a new direction in problem characterisation by exploring programmatic representations of optimisation problem instances. In Section~\ref{sec:litrev}, we review existing ways of characterising problems and discuss the key aspects and limitation of these approaches. In Section 3, we discuss the abstract idea of representation of an optimisation problem and introduce the idea of measuring the complexity of programmatic representations for quantifying similarity/difference between different problem instances. Section 4 proposes the use of two specific measures of program complexity: the Halstead volume and the Shannon entropy, showing that the former can be seen as a special case of the latter. Illustrative examples and implementation details are provided, together with evaluation on the BBOB test set and neural network weight optimisation problems. In Section 5, we show that the measures are generally correlated with algorithm performance. Section 6 provides conclusions and suggestions for future work. Code and data to reproduce the experiments is provided\footnote{Zenodo URL to be provided after acceptance.}.

\section{Problem Classes, Characterisation and Features}
\label{sec:litrev}
An important practical consequence of the No Free Lunch theorems~\cite{wolpert2002no} is the implication that when the space of optimisation problems is restricted in some way, there will be performance differences between some given algorithm instances. This generally motivates work towards identifying algorithms that perform well/poorly for certain classes of problems of interest.

However, the task of defining problem classes is far from simple and a variety of different approaches exist. One possibility is to use any information known from the formulation of the optimisation problem instances. At the highest level, this might include the dimensionality of the problem, the type of decision variables (e.g. discrete or continuous), whether the objective function is noisy or not, the total budget of objective function evaluations and whether the problem is constrained or not. While this provides very little information concerning the structure of the problem landscape, it has been shown that it is still enough to drive successful algorithm selection ``wizards''; notably the ABBO/NGOpt wizard that is a hand-crafted, static selector based purely on this type of information~\cite{trajanov22}. It is also true that the formulation of a problem may deliberatly impose structure on the landscape; e.g. engineering the problem so that the objective function should be convex in order to apply certain classes of algorithms. 

In combinatorial problems, broad problem classes have been named based on the definitional formulation - e.g. graph colouring, travelling salesperson problem, knapsack, bin-packing. Within these classes, ``problem-specific features'' have been used to further categorise problems, e.g. \cite{smith12}. For continuous problems, broad classes can also follow from the problem formulation, by making assumptions such as smoothness or the availability of derivative information.

It is also common to describe optimisation problem classes in a more qualitative way. For example, the five categories of problems in the BBOB noiseless benchmark set~\cite{hansen:BBOB} are: (1) separable, (2) low or moderate conditioning, (3) high conditioning and unimodal, (4) multi-modal with adequate global structure, and (5) multi-modal with weak global structure. Other qualitative terms include ``big valley'' structure~\cite{boese1996models} or ``funnel''~\cite{sutton2006pso}. These terms describe landscape properties in an informal way, but detecting these properties in an unknown landscape is non-trivial. With some effort and/or assumptions, quantitative definitions can sometimes be developed for such terms, however measuring these properties experimentally is a challenge.

Fitness landscape analysis takes a different approach, by modelling and summarising the geometrical or topographical structure of the landscapes of problem instances (as defined by the fitness function, search space and search operators or notion of distance in the search space)~\cite{malan21}. 
Landscape analysis has been shown to be useful in characterising and comparing problem instances and extracted features have been used successfully to build machine learning pipelines for predicting algorithm performance and achieving algorithm selection~\cite{kerschke19b,munoz15}. However, there are important practical limitations and challenges. The popular exploratory landscape analysis (ELA)~\cite{mersmann11} with extended landscape features in the flacco package~\cite{kerschke19} is based on generating and evaluating a sample of solution points over the landscape. It is desirable to keep this sample small since it consumes objective function evaluations that may be expensive or could be used by the optimiser~\cite{kerschke16}. However a small sample is unlikely to be able to provide rich information about the properties of a complex fitness landscape. Previous work has shown that landscape analysis is sensitive to the sample size and sampling strategy used~\cite{munoz2022analyzing,renau20}. Many of the ELA features themselves often estimate a very simple statistic or model of the data points, meaning that they may not be able to distinguish between different problem instances, or present a representation of the instance space that is robust~\cite{gallagher2024towards,vskvorc22}.

Considerable challenges remain when it comes to evaluating the strengths and limitations of landscape features if the goal is to understand or explain the relationship between algorithm performance and landscape features. Alternatively, if landscape features are used as part of a machine learning pipeline for algorithm selection, it seems beneficial to calculate large, diverse feature sets to provide a better characterisation of problem landscapes. Therefore, it is worthwhile exploring a wide variety of alternative representations of problem instances to see if any advantages or benefits can be gained.


\section{Characterising Problems Using Their Programmatic Representation}
In this paper we focus on the continuous unconstrained, single-objective optimisation problem:
\begin{equation}
    \mathbf{x^*} = \arg \min f(\mathbf{x}), x \in \mathcal{X} \subseteq \Re^n
\end{equation}
However, the ideas are quite general and should be applicable to other types of problems (e.g. discrete, constrained, multi-objective).

The structure of an optimisation problem instance is determined by the objective function, $f$. In the context of textbooks and benchmarking, it is natural to think about the objective as a mathematical function that we can write down (even though in black-box optimisation we do not assume any knowledge of the function). However in many optimisation problems, the objective function is implemented partly based on computer simulations, physical measurements or data sets. Evolutionary computation has also been used in many scenarios where the objective function used is a ranking or preference over candidate solutions, which may be determined by human preferences, etc. Along these lines, we take an abstract view that \textit{any} well-defined representation of $f$ could potentially be used to measure properties of the problem, thereby providing useful information for problem characterisation and comparison.

In the following we consider a novel approach based purely on a representation of the objective function as implemented in code (i.e. a computer program in some language). 
The premise is that the complexity inherent in the programmatic implementation of the objective function captures useful information about how it will ultimately manifest as a search landscape. Intuitively, a function that has a simple mathematical structure will tend to be simple to implement in code, needing very few operators and manifests as a smooth landscape. On the other hand, a complex function tends to require a longer description, using many more operators and operands and manifests as a highly irregular landscape.

Considering a programmatic representation opens up the possibility of using ideas from software engineering and information theory to measure the properties of code. As far as we are aware, this is an unexplored avenue of research for optimisation problem instances. Nurminen~\cite{nurminen03} measured the programmatic complexity of shortest path algorithms using Halstead volume, lines of code and cyclomatic measures. The results show a high correlation between these measures and the running time of the different algorithms. While the aims of this work are quite different to ours, the paper makes the general argument that the empirical comparison of algorithms should consider the implementation (in code) of the algorithms to be studied. Information theoretic ideas have been used to develop some landscape features, in particular information content~\cite{vassilev00}, entropy~\cite{malan09} and the adaptation of (normalised) compression distance~\cite{morgan15}. However, as landscape features they are calculated from a sample of data points from the search space and do not consider a programmatic representation.

\section{Measuring Program Complexity of Optimisation Problems }

Measuring the complexity of code is of considerable interest in software engineering. However, much of the work in this area is focussed on the human effort required to develop or debug software. The ``number of lines of code'' is a rough measure of software complexity, but is often used in practice. The programming language used is also a significant factor, since different languages produce code that is more compact or verbose and have a different set of operators and operands. To compare the structural logic of programs, there is the possibility of working with a more general representation, such as a tokenization, abstract syntax tree or any other representation between source code and the executable binary.  

In this paper, our focus is not on aspects related to software development, but rather on the complexity of a block of code that implements a given objective function. Halstead~\cite{halstead77} proposed to measure the properties of a program that is focussed on the representation and structure of the code, in a way that is independent of its execution behaviour on a given architecture. 
In particular, the Halstead volume, $V_H$ is defined as
\begin{equation}
\label{eqn:Hal}
V_H = (N_1+N_2)\log_2(\eta_1+\eta_2),
\end{equation}
where $N_1$ and $N_2$ are the total numbers of operators and operands respectively, and $\eta_1$ and $\eta_2$ the number of distinct operators and operands in a given program. $V_H$ is clearly one measure of the complexity of some entity (e.g. an objective function) expressed as a program. The Halstead difficulty, $D_H$ is defined as
\begin{equation}
D_H = \frac{\eta_1}{2} \times \frac{N_2}{\eta_2},
\end{equation}
which is related to the difficulty of a program to write or understand, as well as the Halstead effort, which may be used to measure required human coding time. We therefore focus on $V_H$ which seems more relevant for our purposes.

The Halstead volume is closely related to the Shannon entropy, also proposed to measure software complexity~\cite{harrison92}. The entropy per token is given by:
\begin{equation}
\label{Eqn:ent}
H_x = - \sum_{x \in \mathcal{X}}p(x)\log_2 p(x)
\end{equation}
where $x$ is a member of the discrete set $\mathcal{X}$ of possible tokens  (operators and operands, i.e. $\eta$) and $p(x)$ is a discrete probability over the total number of tokens in the program, $N$. Harrison~\cite{harrison92} considers only operators (and the total number of operators) and suggests this measure as it is not dependent on the total size of the program. However for code specifically describing objective functions we assume that the total size is relevant. The total entropy of the program is then:
\begin{equation}
\label{Eqn:totent}
H = NH_x
\end{equation}
If we assume uniform probability values for each operator and operand, $H$ becomes equal to $V_H$.

\subsection{Halstead Volume and Entropy: Illustrative Example}
As an example, consider firstly the Sphere function:~~$ f_1(\mathbf{x}) = \sum_{i=1}^n x_i^2$.\\
A simple Python implementation of this function is shown in Listing~\ref{lst:f1py}. For this implementation:
\begin{itemize}
    \item[$\bullet$] $\eta_1 = 10 \; (\texttt{def, (, ), :, =, for, in, +=, *, return})$
    \item[$\bullet$] $\eta_2 =5 \; (\texttt{f1, x, s, 0, v})$
    \item[$\bullet$] $N_1 = 11$
    \item[$\bullet$] $N_2 = 10$
\end{itemize}
so $V_H(f_1) \approx  82.04$. Entropy accounts for the multiple occurrences of tokens \texttt{s, v} ($p(\texttt{s}) = p(\texttt{v}) = \frac{3}{21}$) and \texttt{x,:} ($p(\texttt{x}) = p(\texttt{:}) = \frac{2}{21}$), leading to $H_{f_1} \approx 78.73$ 

Compare this with an implementation of the Lunacek bi-Rastrigin function (BBOB $f_{24}$, Listing~\ref{lst:f24py}), for which we have $V_H(f_{24}) \approx  764.32$ and $H_{f_{24}} \approx 703.24$ ($\eta = 44$, $N = 140$). According to $V_H$, $f_{24}$ is $\approx 9.32$ times more complex than $f_1$ and $\approx 8.93$ times more complex according to $H$. Halstead volume is more sensitive to longer programs with repeated uses of the same tokens, for example nested brackets in a complex mathematical expression.

\begin{listing}[htbp]
\begin{lstlisting}[language=python, caption={Implementation of $f_1$ in Python.}, label={lst:f1py}]
def f1(x):
    s = 0
    for v in x:
        s += v * v
    return s
\end{lstlisting}
\end{listing}

\begin{listing}[htbp]
\begin{lstlisting}[language=python, caption={Implementation of $f_{24}$ in Python.}, label={lst:f24py}]
def f24(x):
    d = len(x)
    s = 1.0 - (1.0 / (2.0 * math.sqrt(d + 20.0) - 8.2))
    mu0 = 2.5
    mu1 = -math.sqrt((mu0**2 - 1.0) / s)
    sum1, sum2, sum3 = 0, 0, 0
    for i in range(d):
        sum1 += (x[i] - mu0)**2
        sum2 += (x[i] - mu1)**2
        sum3 += math.cos(2 * math.pi * (x[i] - mu0))
    res = min(sum1, d + s * sum2) + 10 * (d - sum3)
    return res
\end{lstlisting}
\end{listing}

There are clearly limitations as to what is captured by the programmatic representation. In particular, we can see from the examples that the dimensionality of a problem has no effect on $V_H$ and $H$, since dimensionality is just represented as a single operand and/or the number of times the loop is executed to evaluate the summation in functions such as $f_1$ and $f_{24}$. On the other hand, this limitation is transparent, in comparison to sample-based ELA features that may capture dimensionality to some extent, but the effectiveness depends on the sample size and technique.

\subsection{Practical Considerations for Calculating $V_H$ and $H$}



The examples above suggest the possibility of comparing different optimisation problem instances via Halstead volume and/or entropy. However there are a number of factors to consider before this can be done in practice:
\paragraph{Programming language:} since the measures are dependent on the language used, the best option would be to compare problems based on their implementation in the same language. While the measures are not directly comparable across different languages, the difference could be assumed to be a roughly constant factor, meaning that a ranking of functions according to these measures would be preserved (i.e. if $H(f_1)<H(f_{24})$ for Python implementations, we could expect that $H(f_1)<H(f_{24})$ for C implementations).
\paragraph{Style:} we expect that the style of different programmers would lead to some variability in the measures if these programs were compared. This suggests that the measures are likely to be more reliable when used on programs that were developed by the same programming teams or there is some attempt to adhere to agreed style guidelines or principles. 

\subsection{Complexity Measures of BBOB functions}
To test Halstead volume and entropy for characterising optimisation problems, we perform an analysis on the well known suite of 24 bound-constrained, noiseless BBOB functions~\cite{hansen:BBOB} as they are documented and implemented on the COCO platform~\cite{BBOB-COCO}. We extract only the underlying objective functions as they are implemented in the C language without any wrapper COCO functionality. Error checking code and boundary handling were also removed. For example, the code used for BBOB function $f_1$ is shown in Listing~\ref{lst:f1}. 

To compute the Halstead metrics on the BBOB functions, we firstly used the \texttt{multimetric} Python package\footnote{multimetric 2.4.4 (\url{https://pypi.org/project/multimetric/})}. Multimetric calculates a number of code metrics directly including $V_H$. Internally, it uses the \texttt{Pygments} package\footnote{\url{https://pygments.org/}} which tokenises the code using regular expressions. Alternatively, we also use the Python package for \texttt{Clang}\footnote{ \url{https://pypi.org/project/clang/}} to tokenise the C code, from which we calculate $V_H$ and $H$ using (\ref{eqn:Hal}) and (\ref{Eqn:totent}).  

\begin{listing}[htbp]
\begin{lstlisting}[language=C, caption={BBOB $\textrm{f}_1$ raw function in C.}, label={lst:f1}]
static double f_sphere_raw(const double *x, const size_t number_of_variables) {
  size_t i = 0;
  double result;
  result = 0.0;
  for (i = 0; i < number_of_variables; ++i) {
    result += x[i] * x[i];
  }
  return result;
}
\end{lstlisting}
\end{listing}

\begin{table}[t]
\centering
\caption{Halstead volume ($V_H$) and entropy ($H$) values of base BBOB functions based on the C implementation in COCO.}\label{tab:HV_bbob}
\begin{tabular}{|l|l|r|r|r|}
\hline
Function &  Function Name & $V_H$ (\texttt{multimetric}) & $V_H$ (\texttt{Clang}) & $H$ (\texttt{Clang})\\
\hline
\multicolumn{5}{|c|}{Separable functions}\\\hline
$\textrm{f}_1$ & Sphere & 208.15 & 267.93 & 253.05\\\hline
$\textrm{f}_2$ & Ellipsoid Original&437.47 & 537.67 & 493.42 \\\hline
$\textrm{f}_3$ & Rastrigin Original& 380.80 & 468.05 & 437.22\\\hline
$\textrm{f}_4$ & B\"{u}che-Rastrigin& 394.84 & 482.15 & 451.18\\\hline
$\textrm{f}_5$ & Linear Slope&500.37 & 649.28 & 587.21\\\hline
\multicolumn{5}{|c|}{Functions with low or moderate conditioning}\\\hline
$\textrm{f}_6$ & Attractive Sector& 408.66 & 488.40 & 448.88\\\hline
$\textrm{f}_7$ & Step Ellipsoidal&2434.09 & 2604.00 & 2255.70\\\hline
$\textrm{f}_8$ & Rosenbrock Original& 460.64 & 539.27 & 504.81\\\hline
$\textrm{f}_9$ & Rosenbrock Rotated& 460.64 & 539.27 & 504.81\\\hline
\multicolumn{5}{|c|}{Functions with high conditioning and unimodal}\\\hline
$\textrm{f}_{10}$ & Ellipsoid Rotated& 437.47 & 537.67 & 493.42\\\hline
$\textrm{f}_{11}$ & Discus& 267.93 & 346.13	& 325.61 \\\hline
$\textrm{f}_{12}$ & Bent Cigar& 267.93 & 346.13	& 325.61\\\hline
$\textrm{f}_{13}$ & Sharp Ridge & 509.58 & 615.51 & 571.37\\\hline
$\textrm{f}_{14}$ & Different Powers & 393.50 & 498.55 & 444.07\\\hline
\multicolumn{5}{|c|}{Multi-modal functions with adequate global structure}\\\hline
$\textrm{f}_{15}$ & Rastrigin Rotated&  380.80 & 468.05	& 437.22\\\hline
$\textrm{f}_{16}$ & Weierstrass&  579.03 & 653.53 & 602.45\\\hline
$\textrm{f}_{17}$ & Schaffers F7 & 559.39 & 665.96 & 618.90\\\hline
$\textrm{f}_{18}$ & Schaffers F7 Ill-conditioned& 559.39 & 665.96 & 618.90\\\hline
$\textrm{f}_{19}$ & Griewank-Rosenbrock& 605.33 & 707.20	& 665.35\\\hline
\multicolumn{5}{|c|}{Multi-modal functions with weak global structure}\\\hline
$\textrm{f}_{20}$ & Schwefel&351.86 & 427.35 & 399.38\\\hline
$\textrm{f}_{21}$ & Gallagher 101 Peaks& 1540.77 & 1697.00	& 1512.00\\\hline
$\textrm{f}_{22}$ & Gallagher 21 Peaks & 1540.77 & 1697.00	& 1512.00\\\hline
$\textrm{f}_{23}$ &Katsuura & 766.75 & 906.19 & 837.32\\\hline
$\textrm{f}_{24}$ & Lunacek bi-Rastrigin& 2838.00 & 3024.00 & 2610.70\\\hline
\end{tabular}
\end{table}

Table~\ref{tab:HV_bbob} shows the results for the 24 BBOB functions. We make the following observations:
\begin{itemize}
    \item For all three complexity measures the lowest value is for $\textrm{f}_{1}$ (Sphere) and the highest is for $\textrm{f}_{24}$ (Lunacek bi-Rastrigin), which supports the intuition that higher $V_H$ and $H$ are associated with more complex functions.
    \item The complexity metrics are invariant to rotation. This is evident in functions such as $\textrm{f}_{2}$ (Ellipsoid Original) and $\textrm{f}_{10}$ (Ellipsoid Rotated) having the same $V_H$ and $H$ values.
    \item The complexity metrics are also invariant to different degrees of conditioning (see $\textrm{f}_{17}$ and $\textrm{f}_{18}$).
    \item For the Gallagher Peaks functions ($\textrm{f}_{21}$ and $\textrm{f}_{22}$), the number of peaks is specified as a parameter of the generating function, so is also not captured in the complexity metrics. 
\end{itemize}

\begin{figure}
\includegraphics[width=0.49\textwidth]{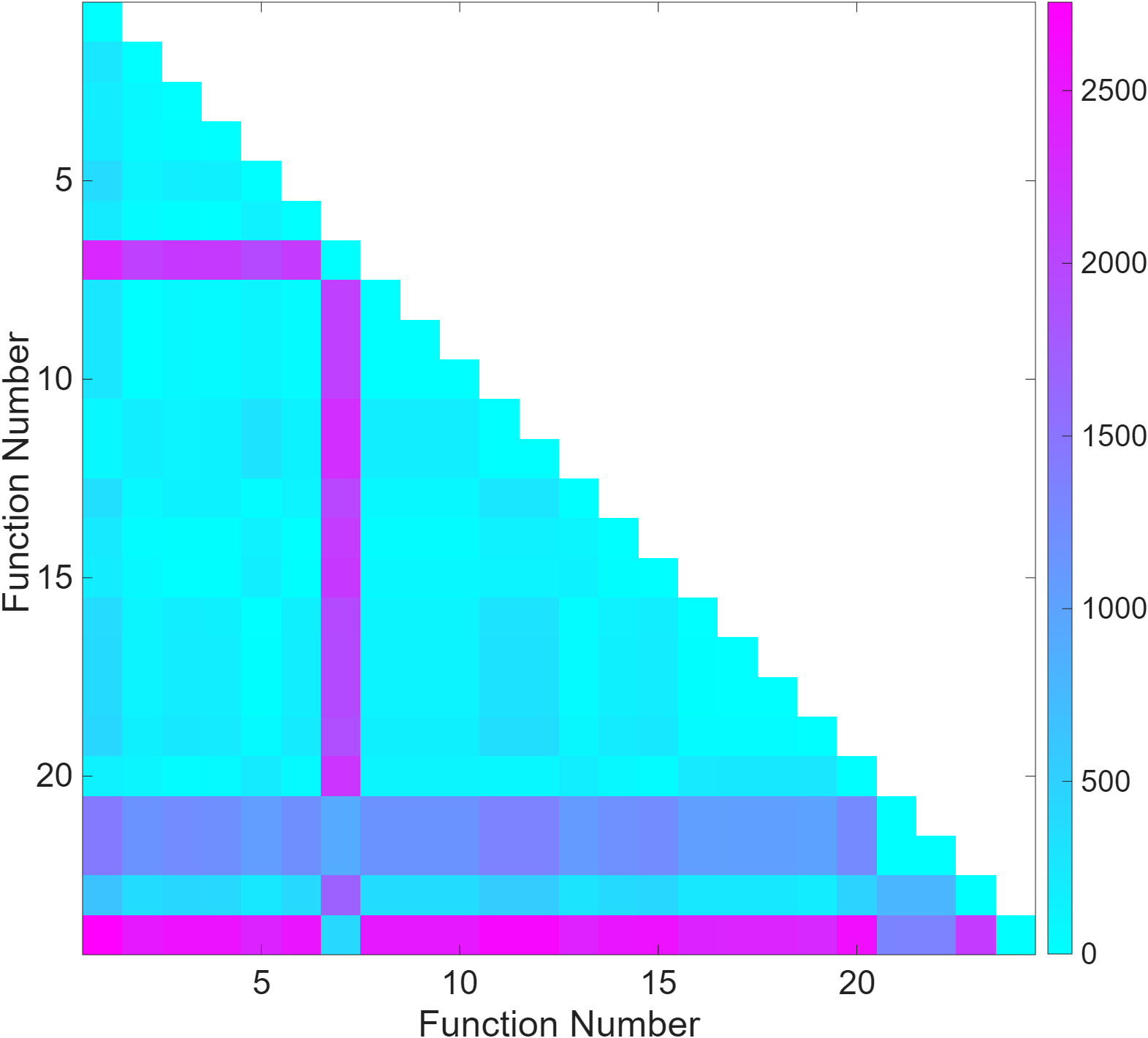}
\includegraphics[width=0.49\textwidth]{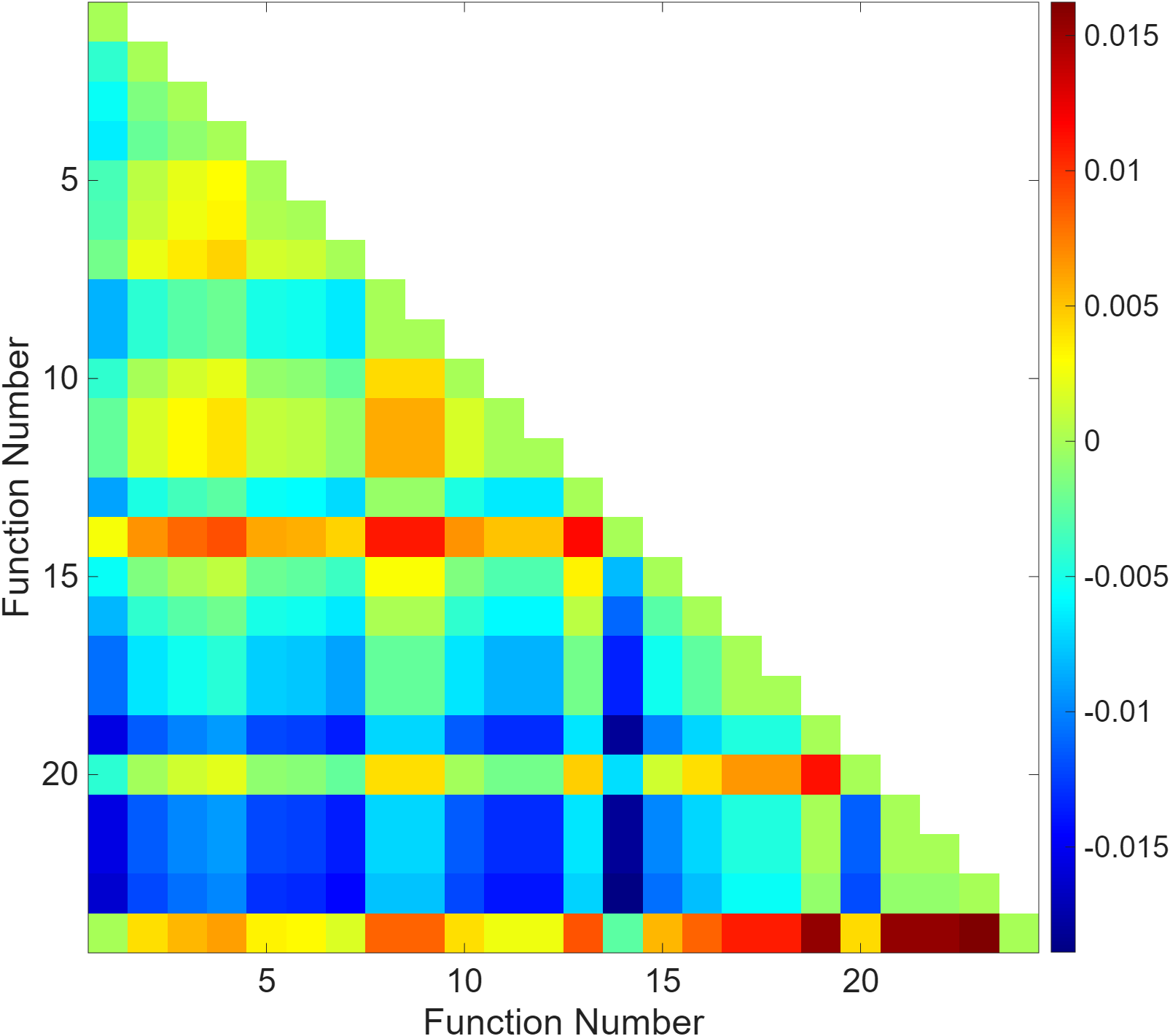}
\caption{BBOB function results. Left: heatmap of differences between $V_H$ values. Right: heatmap of differences between (normalised) $V_H$ and $H$ values.}
\label{fig1}
\end{figure}

Figure~\ref{fig1} (left) shows a heatmap visualisation of the pairwise absolute differences between the $V_H$ (multimetric) values of the BBOB functions (the heatmaps of differences between $V_H$ (Clang) and $H$ (Clang) look very similar, so are not shown). Functions $f_7, f_{21},f_{22}$ and $f_{24}$ clearly stand out because of their relatively large values. To compare with the differences in entropy values, we normalise the values of $V_H$ and $H$ in the range $[0,1]$ and visualise the pairwise difference of $(V_H-H)$ across the functions, see Figure~\ref{fig1} (right). The scale is quite small, indicating that $V_H$ and $H$ are fairly consistent across the problem set. Most of the higher-numbered functions give a higher relative entropy value (resulting in a negative difference), suggesting that the complexity of the functions still produces a relatively uniform distribution over tokens, with $f_{14}, f_{24}$ and to some extent $f_{20}$ being exceptions.

\subsection{Complexity Measures for Neural Network Weight Optimisation}
As a different optimisation problem scenario, we consider the task of searching for the combination of weights that minimises error on a neural network learning task. Consider the training of a simple feed-forward neural network as a black-box optimisation problem (as proposed in the CORNN suite~\cite{Malan2022}). Calculating the objective function involves a forward pass through the network to calculate the output and the corresponding error, which is to be minimised. 

To be comparable with the complexity measures for the BBOB functions, we analysed C implementations for calculating the mean squared error between the output and the target without using any C libraries. We implemented two versions of the code using different activation functions in the hidden layers: rectified linear unit (ReLU) and hyperbolic tangent (Tanh). Note that since the same code is used for different sized networks, the complexity measures are independent of the number of layers and neurons in the network. The complexity measures of these two error functions are provided in Table~\ref{tab:HV_NN}. Due to the higher number of operands and operators required to implement Tanh compared to ReLU, the complexity values are slightly higher.  

\begin{table}
\centering
\caption{Halstead volume ($V_H$) and entropy ($H$) values of neural network mean squared error functions implemented in C without any libraries.}\label{tab:HV_NN}
\begin{tabular}{|l|r|r|r|}
\hline
Activation function & $V_H$ (\texttt{multimetric}) & $V_H$ (\texttt{Clang}) & $H$ (\texttt{Clang}) \\ \hline
ReLU & 931.4 & 1065.3 & 1021.4 \\\hline
Tanh & 1005.3 & 1135.8 & 1103.8\\\hline
\end{tabular}
\end{table}

\begin{figure}[t]
\includegraphics[width=\textwidth]{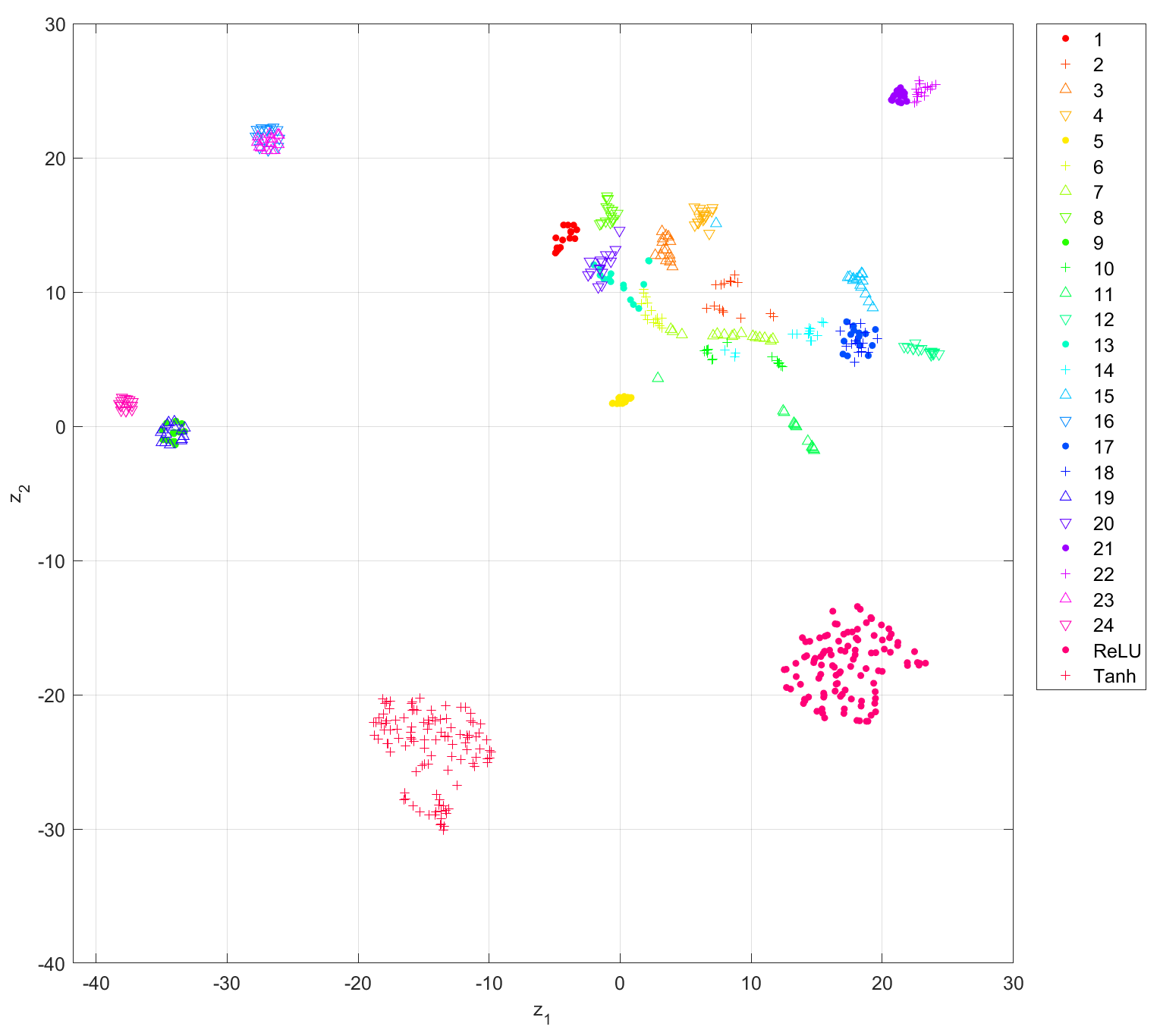}
\caption{2D projection of ELA features of instances of BBOB functions 1-24 and instances of neural network regression tasks from~\cite{Malan2025}.}
\label{fig:tsne}
\end{figure}

The proposed complexity measures consider the objective function on its own for characterising and comparing problems. In the context of neural network weight optimisation, this means that the complexity measures are not affected by the number of weights (the dimension of the search space) or the training dataset (i.e. the learning task). This notion that the characteristics of a neural network optimisation problem are not affected by the learning task was observed in a landscape analysis study of neural network training tasks by Malan and Mu\~{n}oz~\cite{Malan2025}. In that study, they extracted the ELA features of BBOB and neural network regression tasks, equalling the BBOB function dimensions to the number of neural network weights (41 in this case). A 2D projection of the instances (see Fig.~\ref{fig:tsne}) showed that the neural network instances (the two pink clusters at the bottom of the plot) were only differentiated based on the activation function. This was a surprising result, as the neural network training instances covered a wide range of difficulty as regression tasks and yet occupied the same area of ELA feature space. The fact that our proposed complexity measures are only affected by the activation function aligns with the finding that the activation function was the key differentiator between problem instances with respect to landscape features.

\section{Linking Complexity to Performance of Algorithms}
Much of the early work in fitness landscape analysis was focussed on finding a general measure of ``problem hardness'' for evolutionary search~\cite{Malan2013}. Recognising the futility of this endeavour, the focus later shifted to characterise multiple features of problems using a range of different measures with the hope that these feature vectors capture differences between problem instances. If algorithm selection is to be achieved, however, we still need features that in some way link to algorithm performance. 

The ELA feature set is commonly used as the basis for algorithm selection~\cite{kerschke19b}. However, in a large benchmark study, Vermetten et al.~\cite{vermetten2023ma} found that {\it ``the ELA features are not sufficiently representative to accurately represent the problems in a way which is relevant for ranking optimization algorithms''}. In this section, we investigate to what extent Halstead and entropy complexity measures link to algorithm performance and capture aspects of problem difficulty.

\subsection{Performance of Algorithms}
We use the performance data from a benchmark study by Vermetten et al.~\cite{vermetten2023ma} on the BBOB functions in 2 and 5 dimensions. Performance data was shared for five algorithms: (1) Diagonal CMA-ES (dCMA), (2) COBYLA (COB) -- a local gradient-free optimizer, (3) Differential Evolution (DE), (4) Modular CMA-ES (mCMA), and (5) L-SHADE (aDE) -- an adaptive version of DE. Performance was reported as the normalized area under the curve (AUC) of the empirical cumulative distribution function. 

Table \ref{tab:perf} shows the AUC values for the five algorithms for each BBOB function in 2D and 5D. Note that the values displayed are averages of five instances of each BBOB function. We also provide the average AUC ($\overline{\textrm{AUC}}$) of the five algorithms for each function. 
From the AUC values, we notice that they are generally high (close to 1) for $f_1$ and are lower for the later functions, indicating that the algorithms performed better on the early simpler functions, which is as expected. 

\begin{table}
\centering
\caption{AUC performance metrics for five algorithms on BBOB function instances showing negative correlations with Halstead volume.}\label{tab:perf}

\begin{tabular}{|l|c|c|c|c|c|c||c|c|c|c|c|c|}
\hline
&\multicolumn{6}{c||}{2D problem instances} & \multicolumn{6}{c|}{5D problem instances}\\\hline
&dCMA & COB & DE & mCMA & aDE & $\overline{\textrm{AUC}}$ & dCMA & COB & DE & mCMA & aDE & $\overline{\textrm{AUC}}$ \\
\hline
$\textrm{f}_{1}$&0.976 & 0.996 & 0.841 & 0.976 & 0.845 & 0.927 & 0.972 & 0.996 & 0.554 & 0.970 & 0.705 & 0.839 \\
$\textrm{f}_{2}$&0.950 & 0.587 & 0.734 & 0.931 & 0.795 & 0.799 & 0.953 & 0.434 & 0.546 & 0.900 & 0.658 & 0.698 \\
$\textrm{f}_{3}$&0.426 & 0.366 & 0.476 & 0.423 & 0.672 & 0.473 & 0.302 & 0.254 & 0.304 & 0.298 & 0.377 & 0.307 \\
$\textrm{f}_{4}$&0.368 & 0.372 & 0.444 & 0.358 & 0.657 & 0.440 & 0.299 & 0.262 & 0.309 & 0.290 & 0.367 & 0.305 \\
$\textrm{f}_{5}$&0.927 & 0.271 & 0.595 & 0.992 & 0.983 & 0.753 & 0.923 & 0.239 & 0.410 & 0.978 & 0.960 & 0.702 \\
$\textrm{f}_{6}$&0.948 & 0.489 & 0.669 & 0.956 & 0.795 & 0.771 & 0.929 & 0.465 & 0.481 & 0.941 & 0.621 & 0.687 \\
$\textrm{f}_{7}$&0.886 & 0.427 & 0.874 & 0.702 & 0.857 & 0.749 & 0.470 & 0.314 & 0.448 & 0.428 & 0.606 & 0.453 \\
$\textrm{f}_{8}$&0.567 & 0.540 & 0.668 & 0.942 & 0.738 & 0.691 & 0.511 & 0.534 & 0.446 & 0.868 & 0.499 & 0.572 \\
$\textrm{f}_{9}$&0.699 & 0.587 & 0.663 & 0.945 & 0.754 & 0.730 & 0.540 & 0.549 & 0.437 & 0.876 & 0.501 & 0.581 \\
$\textrm{f}_{10}$&0.528 & 0.526 & 0.605 & 0.932 & 0.684 & 0.655 & 0.435 & 0.409 & 0.398 & 0.895 & 0.456 & 0.518 \\
$\textrm{f}_{11}$&0.372 & 0.402 & 0.522 & 0.906 & 0.583 & 0.557 & 0.318 & 0.311 & 0.329 & 0.870 & 0.400 & 0.446 \\
$\textrm{f}_{12}$&0.706 & 0.614 & 0.657 & 0.890 & 0.708 & 0.715 & 0.605 & 0.601 & 0.455 & 0.857 & 0.487 & 0.601 \\
$\textrm{f}_{13}$&0.366 & 0.342 & 0.494 & 0.908 & 0.576 & 0.537 & 0.412 & 0.410 & 0.304 & 0.840 & 0.353 & 0.464 \\
$\textrm{f}_{14}$&0.614 & 0.557 & 0.636 & 0.939 & 0.708 & 0.691 & 0.572 & 0.561 & 0.449 & 0.918 & 0.524 & 0.605 \\
$\textrm{f}_{15}$&0.488 & 0.392 & 0.410 & 0.430 & 0.547 & 0.453 & 0.300 & 0.238 & 0.275 & 0.300 & 0.284 & 0.279 \\
$\textrm{f}_{16}$&0.412 & 0.236 & 0.382 & 0.657 & 0.432 & 0.424 & 0.273 & 0.147 & 0.186 & 0.326 & 0.194 & 0.225 \\
$\textrm{f}_{17}$&0.503 & 0.165 & 0.430 & 0.566 & 0.498 & 0.432 & 0.330 & 0.126 & 0.270 & 0.452 & 0.316 & 0.299 \\
$\textrm{f}_{18}$&0.303 & 0.183 & 0.377 & 0.336 & 0.428 & 0.325 & 0.287 & 0.134 & 0.261 & 0.348 & 0.287 & 0.263 \\
$\textrm{f}_{19}$&0.548 & 0.363 & 0.473 & 0.519 & 0.535 & 0.488 & 0.272 & 0.223 & 0.205 & 0.236 & 0.210 & 0.229 \\
$\textrm{f}_{20}$&0.622 & 0.540 & 0.594 & 0.556 & 0.722 & 0.607 & 0.451 & 0.444 & 0.466 & 0.442 & 0.507 & 0.462 \\
$\textrm{f}_{21}$&0.486 & 0.418 & 0.741 & 0.462 & 0.805 & 0.582 & 0.353 & 0.373 & 0.358 & 0.336 & 0.530 & 0.390 \\
$\textrm{f}_{22}$&0.414 & 0.409 & 0.736 & 0.527 & 0.821 & 0.581 & 0.349 & 0.399 & 0.312 & 0.321 & 0.448 & 0.366 \\
$\textrm{f}_{23}$&0.306 & 0.106 & 0.135 & 0.494 & 0.127 & 0.234 & 0.140 & 0.114 & 0.106 & 0.180 & 0.104 & 0.129 \\
$\textrm{f}_{24}$&0.339 & 0.289 & 0.319 & 0.319 & 0.318 & 0.317 & 0.267 & 0.236 & 0.246 & 0.264 & 0.246 & 0.252 \\
\hline\hline
& -0.370 & -0.531 &	-0.138 & -0.336 & -0.164 & -0.369 & -0.444 &-0.527 & -0.524 & -0.462 &	-0.319 & -0.502
\\ \hline

&\multicolumn{12}{c|}{Spearman's rank correlation with Halstead volume}\\\hline\hline
\end{tabular}
\end{table}

\subsection{Correlation to Algorithm Performance}
Since it is difficult to observe patterns in Table~\ref{tab:perf}, we provide the Spearman's rank correlation coefficients of each column of AUC values with the Halstead volume ($V_H$) given in Table~\ref{tab:HV_bbob} (see bottom two rows of Table~\ref{tab:perf}).

A first observation is that all of the correlation coefficients are negative. This confirms that the Halstead volume is negatively correlated with the AUC for all algorithms. This matches our intuition -- the more complex functions (as measured by $V_H$) are harder to optimise. We also see that there is a range of correlations from weak to moderate. For example, for the DE algorithm on 2D instances, the correlation is weak at $-0.138$, but is stronger on 5D problems at $-0.524$.

\begin{figure}
\includegraphics[width=0.49\textwidth]{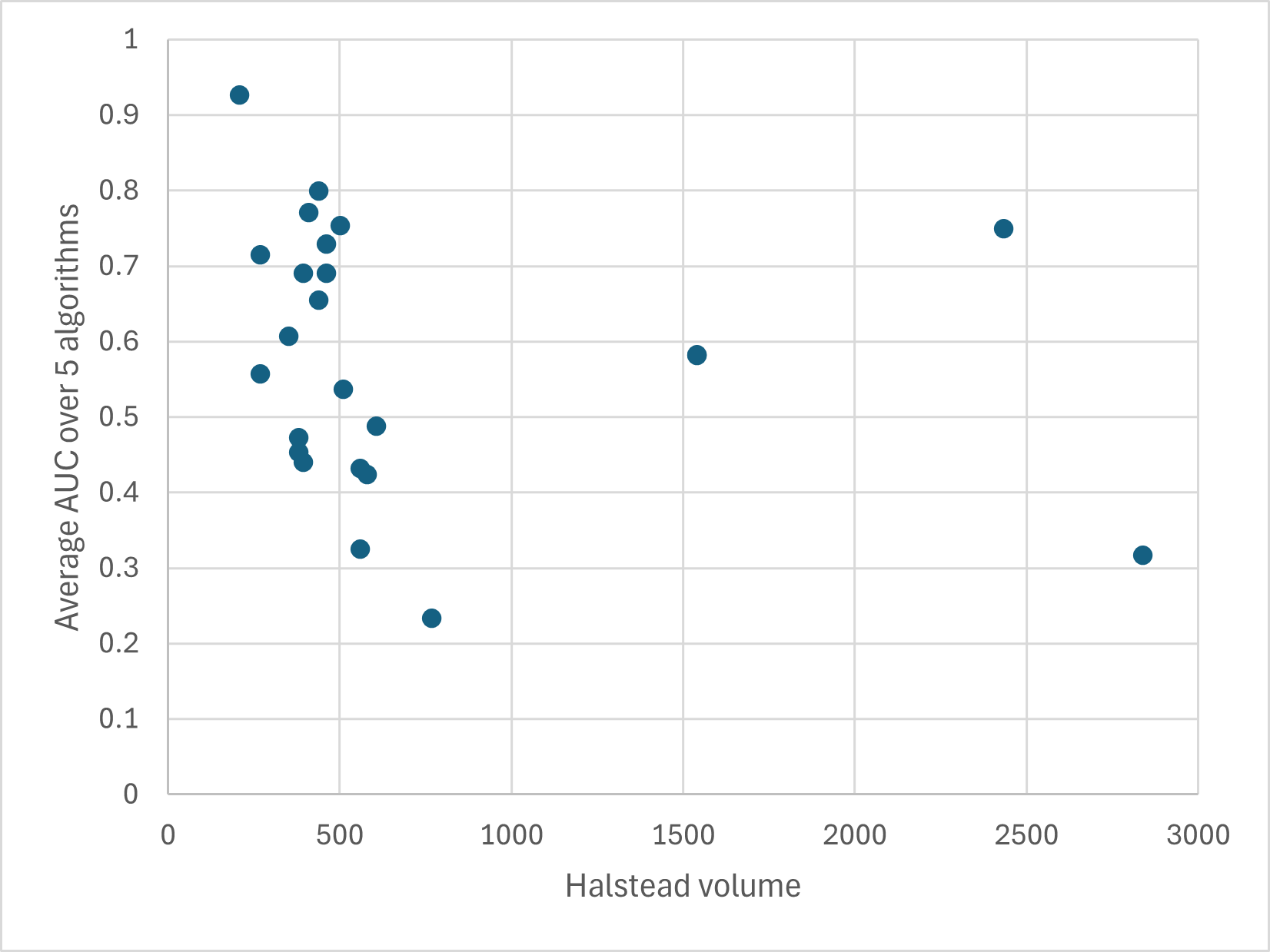}
\includegraphics[width=0.49\textwidth]{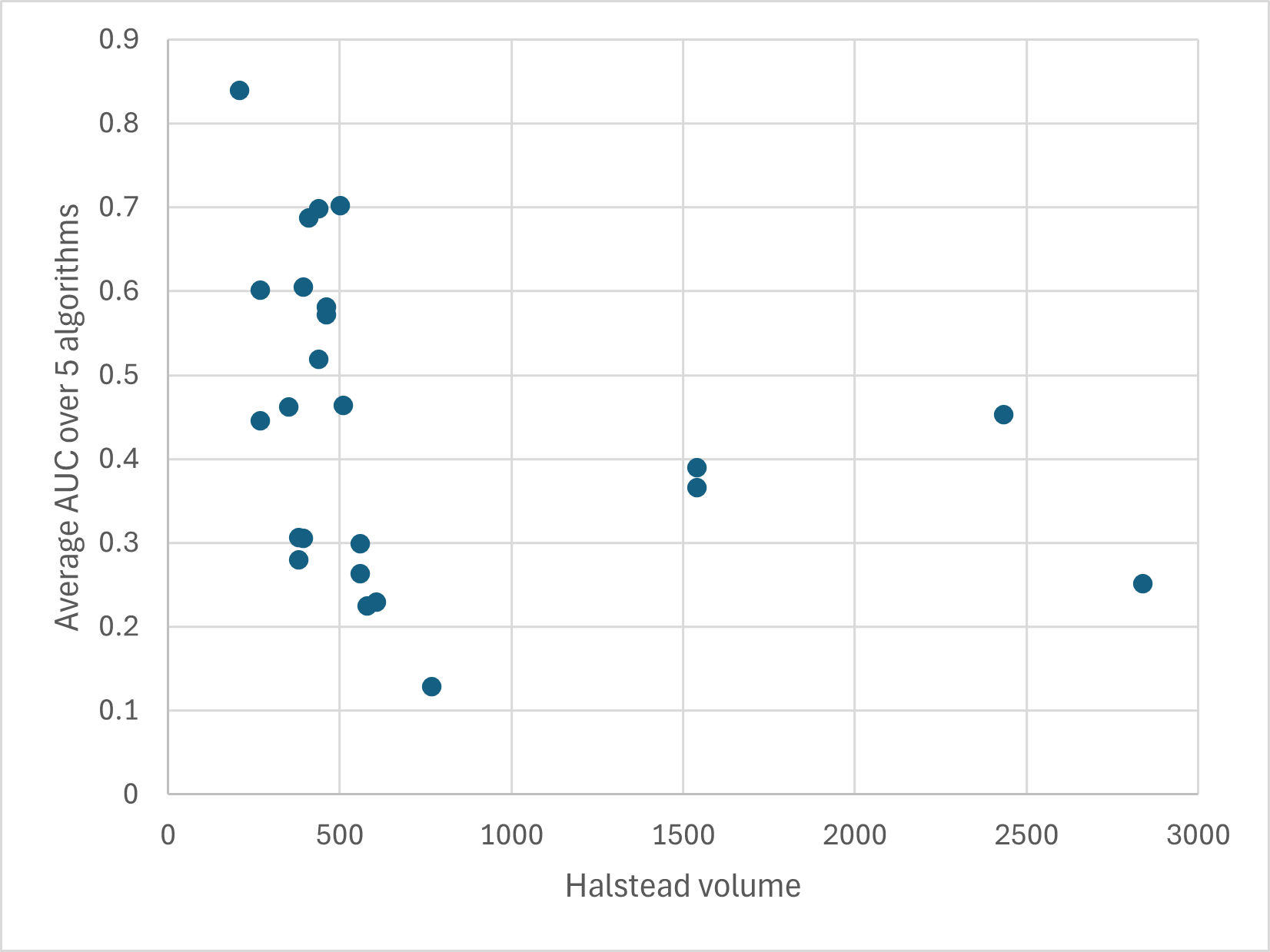}
\caption{Scatter plots of Halstead volume against average AUC for the 2D (left) and 5D (right) BBOB functions.}
\label{fig:AUC_VH}
\end{figure}

To further investigate the correlation, Fig.~\ref{fig:AUC_VH} plots the average AUC (over 5 algorithms) against Halstead volume for the 2D and 5D BBOB functions. A clear trend is shown for $V_H$ values below 1000 in both cases. The four outliers are $f_7$, $f_{21}$, $f_{22}$, and $f_{24}$. This seems to indicate that Halstead volume is a better predictor of algorithm performance when excluding instances with relatively high $V_H$ values.

\section{Conclusion}
In this paper we have proposed the use of programmatic complexity measures as a new approach to characterising and comparing optimisation problem instances. While a programmatic approach cannot capture every property of a problem landscape that might be relevant to algorithm performance, it has a number of benefits. In particular, measuring code complexity does not require sampling, statistical estimation or the use of any objective function evaluations and can be  pre-calculated statically for each problem instance. The Halstead volume and entropy were evaluated and compared experimentally and shown to provide information that could be used to compare problem instances. In addition, the Halstead volume showed correlation to algorithm performance for the BBOB functions, so has potential as a feature to complement existing features in algorithm selection and configuration studies.

There is considerable scope for further work. It would be interesting to see results for a wider set of problems, including real-world representative and problems where the objective function value involves a more extensive execution of code, e.g. a numerical simulation or model. Further, it would be interesting to see if algorithm selectors/wizards can be produced that can take advantage of programmatic complexity to improve performance results. In addition, further work is needed in extend this idea to apply to different classes of optimisation problems, including constrained and multi-objective problems.

\begin{credits}
\subsubsection{\ackname} The authors would like to acknowledge the Dagstuhl Seminar 26072 on ``Best Practice for Leveraging Domain Knowledge in Real-World Optimization'', which provided the opportunity for research discussions that led to the ideas and work in this paper.

\end{credits}
%
%
%
\bibliographystyle{splncs04}
\bibliography{mybibliography}
%




\end{document}